\documentclass[twocolumn,hyperref]{bmcart}
\pdfoutput=1

\usepackage{amsthm,amsmath}
\RequirePackage{hyperref}
\usepackage{latexsym}
\usepackage{enumitem}
\usepackage{graphicx}
\usepackage{tabularx}
\usepackage{multirow}
\usepackage{url}

\newcommand{\CHANGEA}[1]{#1}   

\startlocaldefs
\endlocaldefs

\begin{document}

\begin{frontmatter}

\begin{fmbox}
\dochead{Research}


\title{From POS tagging to dependency parsing for biomedical event extraction}


\author[
   noteref={n1}, 
]{\inits{D}\fnm{Dat Quoc} \snm{Nguyen}}
\author[
   noteref={n2}, 
]{\inits{K}\fnm{Karin} \snm{Verspoor}}



\begin{artnotes}
\note[id=n1]{Correspondence: dqnguyen@unimelb.edu.au, The University of Melbourne }  
\note[id=n2]{Karin.Verspoor@unimelb.edu.au, The University of Melbourne} 
\end{artnotes}



\begin{abstractbox}

\begin{abstract} 

\parttitle{Background} 
Given the importance of relation or event extraction from biomedical research publications to support knowledge capture and synthesis, 
and the strong dependency of approaches to this information extraction task on syntactic information, it is valuable to understand which approaches to syntactic processing of biomedical text have the highest performance.  

\parttitle{Results} 
We perform an empirical study  comparing state-of-the-art traditional feature-based and neural network-based models for two core natural language processing tasks of part-of-speech (POS) tagging and dependency parsing on two benchmark biomedical corpora, GENIA 
and CRAFT. To the best of our knowledge, there  is  no  recent  work  making such comparisons in the biomedical context; specifically no detailed analysis of neural models on this data is available. Experimental results show that in general, the neural models outperform the feature-based models on two benchmark biomedical corpora GENIA and CRAFT. We also perform a task-oriented evaluation to investigate the influences of these models  in a downstream application on biomedical event extraction, and show that better intrinsic parsing performance does not always imply better extrinsic event extraction performance. 

\parttitle{Conclusion} We have presented a detailed empirical study comparing   traditional feature-based and neural network-based models for POS tagging and dependency parsing in the biomedical context, and  also investigated the influence of parser selection for a biomedical event extraction downstream task.

\parttitle{Availability of data and material} We make the retrained models available at \url{https://github.com/datquocnguyen/BioPosDep}.

\end{abstract}




\end{abstractbox}
\end{fmbox}

\end{frontmatter}




\section*{Background}
The biomedical literature, as captured in the parallel  repositories of PubMed\footnote{{\url{https://www.ncbi.nlm.nih.gov/pubmed}}}  (abstracts) and PubMed Central\footnote{{\url{https://www.ncbi.nlm.nih.gov/pmc}}} (full text articles), is growing at a remarkable rate of over one million publications per year. Effort to catalog the key research results in these publications demands  automation~\cite{Baumgartner2007}. Hence extraction of relations and events from the published literature has become a key focus of the biomedical natural language processing community.

Methods for information extraction typically make use of linguistic information, with a specific emphasis on the value of  dependency parses. A number of linguistically-annotated resources, notably including the GENIA \cite{I05-2038} and CRAFT \cite{Verspoor2012} corpora, have been produced to support development and evaluation of natural language processing (NLP) tools over biomedical publications, based on the observation of the substantive differences between these domain texts and general English texts, as captured in resources such as the Penn Treebank~\cite{Marcus93building} that are standardly used for development and evaluation of syntactic processing tools. Recent work on biomedical relation extraction 
has highlighted the particular importance of syntactic information \cite{TACL1028}. Despite this, that work, and most other related work, has simply adopted a tool to analyze the syntactic characteristics of the biomedical texts without consideration of the appropriateness of the tool for these texts. A commonly used tool is the Stanford CoreNLP dependency parser \cite{D14-1082}, although domain-adapted parsers (e.g.  \cite{mcclosky2008self}) are  sometimes used.

Prior work on the CRAFT treebank demonstrated substantial variation in the performance of syntactic processing tools for that data \cite{Verspoor2012}.
Given the significant improvements in parsing performance in the last few years, thanks to renewed attention to the problem and exploration of neural methods, it is important to revisit whether the commonly used tools remain the best choices for syntactic analysis of biomedical texts. In this paper, we therefore investigate current state-of-the-art (SOTA) approaches to dependency parsing as applied to biomedical texts. We also present detailed results on the precursor task of POS tagging, since parsing depends heavily on POS tags. 
Finally, we study the impact of parser choice on biomedical event extraction, following the structure of the extrinsic parser evaluation shared task (EPE 2017) for biomedical event extraction   \cite{epe2017bio}. We find that differences in overall intrinsic parser performance do not consistently explain differences in information extraction performance.

\section*{Experimental methodology}

\begin{figure}[t]
    \centering
    \includegraphics[width=7.5cm]{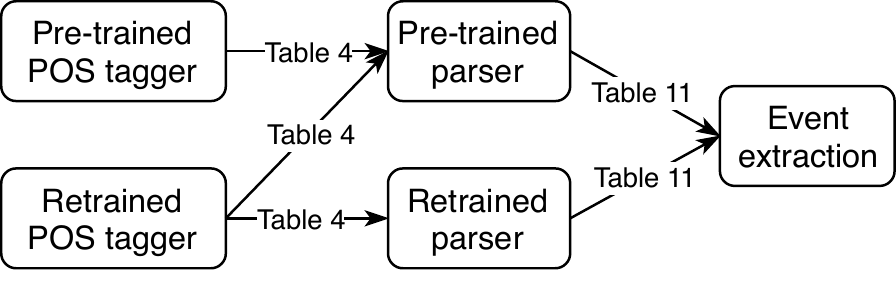}
    \caption{\CHANGEA{Diagram outlining the design of experiments.}}
    \label{fig:Diagram}
\end{figure}

In this section, we present our empirical approach to evaluate different POS tagging and dependency parsing models  on benchmark biomedical corpora. \CHANGEA{Figure \ref{fig:Diagram} illustrates our experimental flow. In particular, we compare pre-trained and retrained POS taggers, and  investigate the effect of these pre-trained and retrained taggers in pre-trained parsing models (in the first five rows of Table \ref{tab:mainresults}). We then compare the performance of retrained parsing models to the pre-trained ones (in the last ten rows of Table \ref{tab:mainresults}). Finally, 
we investigate the influence of pre-trained and retrained parsing models in the biomedical   event extraction task (in Table \ref{tab:bionlp}).}

\subsection*{Datasets}

We use two biomedical corpora: GENIA \cite{I05-2038} and CRAFT \cite{Verspoor2012}. GENIA includes abstracts from PubMed, while CRAFT includes full text publications. It has been observed that there are substantial linguistic differences between the abstracts and the corresponding full text publications
\cite{cohen2010structural}; hence it is important to consider both contexts when assessing  NLP tools in biomedical domain.

The GENIA corpus contains 18K sentences ({$\sim$}486K words) from 1,999 Medline abstracts, which are manually annotated following the Penn Treebank (PTB) bracketing guidelines \cite{I05-2038}.  
On this  treebank, we use the training, development and test split from \cite{david2010}.\footnote{{\url{https://nlp.stanford.edu/~mcclosky/biomedical.html}}}   
We then use  the Stanford constituent-to-dependency conversion toolkit (v3.5.1) to  generate dependency trees with basic Stanford dependencies \cite{deMarneffe:2008:STD:1608858.1608859}.   

The CRAFT corpus includes 21K sentences ({$\sim$}561K words) from 67 full-text biomedical journal articles.\footnote{\url{http://bionlp-corpora.sourceforge.net/CRAFT}} These sentences are   syntactically annotated using an extended PTB tag set.
Given this extended set, the Stanford conversion toolkit is not  suitable for generating dependency trees. Hence,
a dependency treebank using the CoNLL 2008 dependencies \cite{surdeanu-EtAl:2008:CONLL} was produced from the CRAFT treebank using ClearNLP \cite{Choi2012}; we directly use this dependency treebank in our experiments. We use sentences from the first 6 files (PubMed IDs: 11532192--12585968) for development and sentences from the next 6  files (PubMed IDs: 12925238--15005800) for testing, while the the remaining 55 files are used for training.   

Table  \ref{tab:datasets} gives an overview of the experimental datasets, while  Table  \ref{tab:statistics} details corpus statistics.  \CHANGEA{We also include out-of-vocabulary (OOV)  rate in Table \ref{tab:datasets}. OOV rate is relevant because if a
word has not been observed in the training data at all, the tagger/parser is limited to
using contextual clues to resolve the label (i.e.\ it has observed no prior usage
of the word during training and hence has no experience with the word to
draw on).}

\begin{table}[!t]
\caption{\CHANGEA{The number of files (\#file), sentences (\#sent),  word tokens (\#token) and out-of-vocabulary (OOV) percentage in each experimental dataset.}}
\centering
\def\arraystretch{1.05}
\begin{tabular}{ll|llll}
\hline
 \multicolumn{2}{c|}{\bf Dataset} & \textbf{\#file} & \textbf{\#sent} & \textbf{\#token} & \textbf{OOV} \\
\hline 
\multirow{3}{*}{\rotatebox[origin=c]{90}{GENIA}} & Training & 1,701 &  15,820 & 414,608 & 0.0 \\
& Development & 148 &  1,361  & 36,180 & 4.4 \\
& Test & 150 & 1,360 & 35,639 & 4.4 \\
\hline
\multirow{3}{*}{\rotatebox[origin=c]{90}{CRAFT}} & Training & 55 & 18,644 & 481,247 & 0.0\\
& Development & 6 & 1,280 & 31,820 & 6.6 \\
& Test & 6 & 1,786 & 47,926 & 6.3 \\
\hline
\end{tabular}
\label{tab:datasets}
\end{table}

\begin{table*}[!t]
\centering
\caption{Statistics   by the most frequent dependency and overlapped POS labels, sentence length (i.e.\ number of words in the sentence) and relative dependency distances $i-j$ from a dependent $w_i$ to its head $w_j$. In addition,  \%$_{G}$ and  \%$_{C}$ denote the occurrence proportions in GENIA and CRAFT, respectively.} 
\def\arraystretch{1.05}
\begin{tabular}{ll|ll|lll|ll|rll}
\hline
 \multicolumn{4}{c|}{\bf Dependency labels} &  \multicolumn{3}{c|}{\multirow{2}{*}{\bf POS tags}} &  \multicolumn{2}{c|}{\multirow{2}{*}{\bf Length}} &  \multicolumn{3}{c}{\multirow{2}{*}{\bf Distance}}\\
\cline{1-4}
\multicolumn{2}{c|}{GENIA} & \multicolumn{2}{c|}{CRAFT} & & &  & & & &  \\
\hline
Type & \%  &   Type & \%  & Type & \%$_{G}$ & \%$_{C}$ & Type & \% & Type & \%$_{G}$ & \%$_{C}$    \\
\hline
advmod & 2.3 &  ADV & 4.0 & CC & 3.6  & 3.2 & \multicolumn{2}{c|}{\underline{GENIA}}  &  ${<}$ $-5$ & 4.1 & 3.9 \\
amod & 9.6 &  AMOD & 1.9 &CD & 1.6 & 4.0 & 1-10 & 3.5 &  $-5$ &  1.2 & 1.2 \\ 
appos & 1.2 & CONJ & 3.6 &DT & 7.6 & 6.6 &11-20 & 31.0 &  $-4$ & 2.1 & 2.1\\ 
aux & 1.4 & COORD & 3.2 &IN & 12.9 & 11.3  &21-30 & 35.7 & $-3$ & 4.4 & 3.2  \\ 
auxpass & 1.5 &DEP & 1.0 &JJ & 10.1 & 7.6  & 31-40 & 19.4 &  $-2$ & 10.6 & 8.5  \\ 
cc & 3.5 &LOC & 1.7 &NN & 29.3& 24.2  &41-50 & 7.1 & $-1$ &  24.1 &  21.7 \\ 
conj & 3.9 &NMOD & 33.7 &NNS & 6.9 & 6.6 &  $>$ 50 & 3.3 &  1 &  19.0 &  26.5 \\
dep & 2.1 &OBJ & 2.8 &RB & 2.5 & 2.4 &  &   & 2 & 9.4 & 9.8  \\
det & 7.2 & P & 18.4 &TO & 1.6 & 0.6  & \multicolumn{2}{c|}{\underline{CRAFT}}  & 3 & 6.3  &  5.9 \\
dobj & 3.1 &PMOD & 10.6 &VB & 1.1  & 1.1 &1-10 & 17.8 & 4 & 4.0  & 3.4  \\
mark & 1.1 &  PRD & 0.9 & VBD & 2.1& 2.2 & 11-20 & 23.1 &  5 &2.4  & 2.3 \\
nn & 11.6 &PRN & 1.9 &VBG & 1.0& 1.1 &  21-30 & 25.2 & $>$ 5 & 12.3  & 11.6  \\ 
nsubj & 4.1 &ROOT & 3.9 &VBN & 3.1& 3.8 &31-40 & 17.5 & - & -  & - \\ 
nsubjpass & 1.4 &SBJ & 4.9 &VBP & 1.4  & 1.1 &41-50 & 9.3 &  - & -  & -\\ 
num & 1.2 &SUB & 0.9 & VBZ & 1.9  & 1.4 &  $>$ 50 & 7.1 & - & -  & - \\ 
pobj & 12.2 &TMP & 0.9 & - & - & -& - & - & - & -  & - \\
prep & 12.3 &VC & 2.4 & - & - & - & - & - & - & -  & - \\ 
punct & 10.4 & - & - &  - & - & - & - & -& - & -  & - \\ 
root & 3.8 & - & - &  - & - & - & - & -& - & -  & - \\ 
\hline
\end{tabular}
\label{tab:statistics}
\end{table*}

\subsection*{POS tagging models}

We compare SOTA feature-based
and neural network-based models for POS tagging over both GENIA and CRAFT. We consider the following:

\begin{itemize}

	\item \textbf{MarMoT}    \cite{mueller-schmid-schutze:2013:EMNLP} is a well-known generic CRF framework as well as a leading POS and morphological tagger.\footnote{{\url{http://cistern.cis.lmu.de/marmot}}}

	\item {NLP4J}'s   POS  tagging model \cite{choi:2016:N16-1} (\textbf{NLP4J-POS}) is a dynamic feature induction model that automatically optimizes feature combinations.\footnote{{\url{https://emorynlp.github.io/nlp4j/components/part-of-speech-tagging.html}}} NLP4J is the successor of ClearNLP.

    \item \textbf{BiLSTM-CRF}  \cite{HuangXY15} is a sequence labeling model which extends a standard BiLSTM neural network \cite{Schuster1997BRN,HochreiterSchmidhuber1997b} with a CRF layer \cite{Lafferty:2001}.
    
    \item \textbf{BiLSTM-CRF+CNN-char} extends the  model BiLSTM-CRF with character-level word embeddings. For each word token, its  character-level word embedding is derived by applying a CNN to the word's character sequence  \cite{ma-hovy:2016:P16-1}. 
    
    \item \textbf{BiLSTM-CRF+LSTM-char}   also extends the BiLSTM-CRF model with character-level word embeddings, which are derived by applying a BiLSTM to each word's character sequence \cite{lample-EtAl:2016:N16-1}.
    
\end{itemize}

For the three BiLSTM-CRF-based sequence labeling models, we use a performance-optimized implementation from \cite{reimers-gurevych:2017:EMNLP2017}.\footnote{\url{https://github.com/UKPLab/emnlp2017-bilstm-cnn-crf}}
As detailed later in the POS tagging results section,  
we use NLP4J-POS to predict POS tags on development and test sets and perform 20-way jackknifing \cite{koo-carreras-collins:2008:ACLMain} to generate POS tags on the training set for dependency parsing.  

\subsection*{Dependency parsers}

Our second study assesses the performance of SOTA dependency parsers, as well as commonly used parsers, on biomedical texts. 
Prior work on the CRAFT treebank identified the domain-retrained ClearParser~\cite{choi2009k}, now part of the NLP4J toolkit~\cite{choi-tetreault-stent:2015:ACL-IJCNLP}, as a top-performing system for dependency parsing over that data. It remains the best performing non-neural model for dependency parsing.
 In particular, we compare the following  parsers:

\begin{itemize} 

\item  The Stanford neural network dependency parser  \cite{D14-1082}  (\textbf{Stanford-NNdep}) is a greedy  transition-based  parsing model   which concatenates word, POS tag and arc label embeddings into a single vector, and then feeds this vector into 
a multi-layer perceptron with one hidden layer 
for transition classification.\footnote{{\url{https://nlp.stanford.edu/software/nndep.shtml}}}

\item {NLP4J}'s dependency parsing model  \cite{choi-mccallum:2013:ACL2013} (\textbf{NLP4J-dep})    is  a transition-based parser with a selectional branching method  that uses confidence estimates to decide when employing a beam.\footnote{{\url{https://emorynlp.github.io/nlp4j/components/dependency-parsing.html}} }

\item \textbf{jPTDP} v1   \cite{nguyen-dras-johnson-2017} is a  joint model for POS tagging and dependency parsing,\footnote{{\url{https://github.com/datquocnguyen/jPTDP}}} which uses BiLSTMs to learn feature representations shared between POS tagging and dependency parsing. jPTDP can be viewed as an extension of the  graph-based dependency parser \textsc{bmstparser} \cite{TACL885}, replacing POS tag embeddings with LSTM-based character-level word embeddings. For jPTDP, we train with gold standard POS tags.
 
\item The  Stanford ``\textbf{Biaffine}'' parser v1  \cite{DozatM16} extends \textsc{bmstparser} with biaffine classifiers to predict dependency arcs and labels, obtaining the highest parsing result to date on the benchmark English PTB.  
The Stanford Biaffine  parser v2 \cite{dozat-qi-manning:2017:K17-3}, further extends  v1 with  LSTM-based character-level word embeddings, obtaining the highest result (i.e., $1^{st}$ place) at the CoNLL 2017 shared task on multilingual dependency parsing \cite{zeman-EtAl:2017:K17-3}.   We use  the Stanford Biaffine  parser v2 in our experiments.\footnote{{\url{https://github.com/tdozat/Parser-v2}}}
 
\end{itemize}

\subsection*{Implementation details}

We use the training set  to learn model parameters while we tune the model hyper-parameters on the development set. Then we report final evaluation results on the test set. The metric for POS tagging is the accuracy. The metrics for dependency parsing are the labeled attachment score (LAS) and unlabeled attachment score (UAS): LAS is the proportion of words which are correctly
assigned both dependency arc and label
while UAS is the proportion of words for which
the dependency arc is assigned correctly.

For  the three BiLSTM-CRF-based models, Stanford-NNdep, jPTDP and Stanford-Biaffine which utilizes pre-trained word embeddings,  we employ 200-dimensional pre-trained word vectors from \cite{chiu-EtAl:2016:BioNLP16}. 
These pre-trained vectors were obtained by training the Word2Vec skip-gram model \cite{Mikolov:2013} on a PubMed abstract corpus of 3 billion word tokens.

For the traditional feature-based models MarMoT, NLP4J-POS and NLP4J-dep, we use their original pure Java implementations with default hyper-parameter settings. 

For the BiLSTM-CRF-based models, we use default hyper-parameters provided in \cite{reimers-gurevych:2017:EMNLP2017}  with the following exceptions: for training, we use Nadam \cite{Dozat2015IncorporatingNM} and run for 50 epochs. We perform a grid search of hyper-parameters  to select the number of BiLSTM layers from $\{1, 2\}$ and the number of LSTM units in each layer from \{100, 150, 200, 250, 300\}.  Early stopping is applied when no performance improvement on the development set is obtained after 10 contiguous epochs.

For Stanford-NNdep, we select the  \textsf{\small{wordCutOff}}  from $\{1, 2\}$ and  the size of the hidden layer  from \{100, 150, 200, 250, 300, 350, 400\} and fix other hyper-parameters  with their default values. 
 
 For jPTDP, we use 50-dimensional character embeddings and fix the initial learning rate at 0.0005. We also fix the number of BiLSTM layers  at 2 and select the number of LSTM units in each layer from $\{100, 150, 200, 250, 300\}$. Other hyper-parameters  are set at their default values. 

For Stanford-Biaffine, we use default hyper-parameter values  \cite{dozat-qi-manning:2017:K17-3}. These default values can be considered as optimal ones as they helped producing the highest scores for 57 test sets (including English test sets) and second highest scores for  14 test sets over total 81 test sets across 45 different languages at the CoNLL 2017 shared task  \cite{zeman-EtAl:2017:K17-3}.

\section*{Main results}

\subsection*{POS tagging results}
\label{ssec:pos}

Table \ref{tab:POSresults} presents POS tagging accuracy of each model on the test set, based on retraining of the POS tagging models on each biomedical corpus.  
The penultimate row  presents the result of the pre-trained Stanford POS tagging model \textsf{{english-bidirectional-distsim.tagger}} \cite{Toutanova:2003:FPT:1073445.1073478}, 
 trained  on a larger corpus of sections 0--18 (about 38K sentences) of English PTB WSJ text; 
given the use of newswire training data, \CHANGEA{it is unsurprising} that this model produces lower accuracy than the retrained  tagging  models.  
The final row includes published results of the 
GENIA POS tagger \cite{11573036_36}, when trained on  90\% of the GENIA corpus (cf.\ our 85\% training set).\footnote{Trained on the PTB sections 0--18, the accuracies for the GENIA tagger,   Stanford  tagger, MarMoT, NLP4J-POS, BiLSTM-CRF and  BiLSTM-CRF+CNN-char   on the benchmark test set of   PTB  sections 22-24  were reported at 97.05\%,  97.23\%, 97.28\%, 97.64\%, 97.45\% and 97.55\%, respectively.}    It does not support a (re)-training process.

 \begin{table}[!t]
 \caption{POS tagging accuracies on the test set  with gold tokenization. [$\star$]  denotes a result with a pre-trained  POS tagger. \CHANGEA{We do not provide accuracy results of the pre-trained  POS taggers on CRAFT because CRAFT uses an extended PTB POS tag set (i.e.\ there are POS tags in CRAFT that are not defined in the original PTB POS tag set). Corpus-level accuracy differences of at least 0.17\% in GENIA and 0.26\% in CRAFT between two POS tagging models are significant at $p \leq 0.05$. Here, we  compute sentence-level accuracies, then use paired t-test to measure the significance level.}}
\centering
\def\arraystretch{1.05}
\begin{tabular}{l|l|l}
\hline
\textbf{Model} & \textbf{GENIA} & \textbf{CRAFT} \\
\hline 
MarMoT & 98.61 &	97.07 \\
 jPTDP-v1 &  98.66	& 97.24 \\
NLP4J-POS  & 98.80 & 97.43  \\
BiLSTM-CRF &  98.44	& 97.25   \\
 \ \ \ \ \ + CNN-char &  \textbf{98.89}	& 97.51  \\
 \ \ \ \ \ + LSTM-char & 98.85	 & \textbf{97.56}   \\
\hline
Stanford tagger [$\star$] & 98.37 & \_ \\
GENIA tagger [$\star$] & 98.49 & \_ \\
\hline
\end{tabular}
\label{tab:POSresults}
\end{table}

\begin{table*}[t]
\centering
\caption{Parsing results on the test set with predicted POS tags and gold tokenization \CHANGEA{(except [$\mathcal{G}$] which denotes results when employing gold POS tags  in both training and testing phases).} ``Without punctuation''  refers to results excluding  punctuation and other symbols from evaluation.  ``Exact match'' denotes the percentage of sentences whose predicted trees are entirely correct \cite{choi-tetreault-stent:2015:ACL-IJCNLP}. 
[$\bullet$] denotes the use of the pre-trained Stanford  tagger for predicting POS tags  on test set, instead of using the retrained NLP4J-POS model. 
Score differences between the ``retrained''  parsers on both corpora are  significant at $p \leq 0.001$ using  McNemar's test  \CHANGEA{(except  UAS scores obtained by Stanford-Biaffine-v2 for gold and predicted POS tags  on GENIA, i.e.\ 92.51 vs. 92.31 and 92.84 vs. 92.64, where $p \leq 0.05$).}
}
\def\arraystretch{1.05}
\begin{tabular}{lll|ll|ll|ll|ll}
\hline
\multicolumn{3}{c|}{\multirow{3}{*}{\bf System}} & \multicolumn{4}{c|}{\bf With punctuation} & \multicolumn{4}{c}{\bf Without punctuation}\\
\cline{4-11}
& & & \multicolumn{2}{c|}{\bf Overall} & \multicolumn{2}{c|}{\bf Exact match} & \multicolumn{2}{c|}{\bf Overall} & \multicolumn{2}{c}{\bf Exact match}\\
\cline{4-11}
\cline{4-11}
& & &   LAS & UAS  &   LAS & UAS  &   LAS & UAS &   LAS & UAS \\
\hline

\multirow{5}{*}{\rotatebox[origin=c]{90}{GENIA}} & \multirow{5}{*}{\rotatebox[origin=c]{90}{\textbf{Pre-trained}}} 
& Stanford-NNdep [$\bullet$]  & 86.66	& 88.22  & 25.15	 & 29.26	 & 87.31	& 89.02  & 25.88 & 	30.22 \\
& & Stanford-Biaffine-v1  [$\bullet$]   & 84.69	& 87.95	 & 16.25& 	26.10 & 	  84.92	& 88.55	  & 16.99	& 28.24	 \\
& & Stanford-NNdep  & 86.79 & 	88.13  & 25.22 & 29.19  & 87.43 & 	88.91  & 25.88 &	30.15 \\
& & Stanford-Biaffine-v1  & 84.72 &	87.89 & 16.47 &	25.81	 & 84.94 &	88.45 &	 17.06 &	27.79	 \\
& & BLLIP+Bio & \textbf{88.38} & \textbf{89.92} & \textbf{28.82}	& \textbf{35.96}  & \textbf{88.76}	& \textbf{90.49} & \textbf{29.93}	& \textbf{37.43} \\
\hline
\multirow{5}{*}{\rotatebox[origin=c]{90}{GENIA}} &   \multirow{5}{*}{\rotatebox[origin=c]{90}{\textbf{Retrained}}} & Stanford-NNdep  &  87.02	 & 88.34 & 25.74	& 30.07 & 87.56 &	89.02 & 26.03 & 	30.59\\
& & NLP4J-dep 	 &  88.20 &	89.45 & 28.16	& 31.99 & 88.87	& 90.25 & 28.90	& 32.94 \\
& & jPTDP-v1  &  90.01 &	91.46 & 29.63	& 35.74 & 90.27 & 	91.89 &30.29 &	37.06 \\
& &  Stanford-Biaffine-v2   & \textbf{91.04}	& \textbf{92.31} & \textbf{33.38} & 	\textbf{39.56} & \textbf{91.23}	& \textbf{92.64} &\textbf{34.41} &	\textbf{41.10}\\
& &  Stanford-Biaffine-v2 [$\mathcal{G}$]  & 91.68 &	92.51 & 36.99 &	40.44 & 91.92	& 92.84 & 38.01	& 41.84 \\
\hline
\multirow{5}{*}{\rotatebox[origin=c]{90}{CRAFT}} &   \multirow{5}{*}{\rotatebox[origin=c]{90}{\textbf{Retrained}}} & Stanford-NNdep  &  84.76	& 86.64 & 25.31 & 30.40 & 85.59	 & 87.81 & 25.48 &	30.96\\
& & NLP4J-dep 	 &  86.98	& 88.85 & 27.60 &	33.71 & 87.62	& 89.80 & 28.16	& 34.60\\
& & jPTDP-v1  &  88.27	& 90.08 & 29.68	& 36.06 & 88.66 &	90.79 & 30.24	& 37.12\\
& &  Stanford-Biaffine-v2   &  \textbf{90.41} &	\textbf{92.02} & \textbf{33.20}	& \textbf{40.03} & \textbf{90.77}	& \textbf{92.67} & \textbf{33.87}	 & \textbf{41.10}\\
& &  Stanford-Biaffine-v2  [$\mathcal{G}$] & 91.43	& 92.93 & 35.22 &	41.99 & 91.69	& 93.47 & 35.61	& 42.95 \\ 
\hline
\end{tabular}
\label{tab:mainresults}
\end{table*}

 In general, we find that the six retrained  models produce competitive results. 
 BiLSTM-CRF and MarMoT  obtain  the lowest scores on GENIA and CRAFT, respectively. jPTDP obtains a similar score to MarMoT on GENIA and similar score to BiLSTM-CRF on CRAFT.   
In particular, MarMoT obtains accuracy results at 98.61\% and 97.07\% on GENIA and CRAFT, which are  about 0.2\% and 0.4\% absolute lower than   NLP4J-POS, respectively.  
NLP4J-POS uses additional features based on Brown clusters \cite{Brown:1992} and pre-trained word vectors learned from a large external corpus, providing useful extra information. 
 
 BiLSTM-CRF  obtains accuracies of 98.44\% on GENIA and 97.25\% on CRAFT. Using character-level word embeddings helps to produce about 0.5\% and 0.3\% absolute improvements to  BiLSTM-CRF on GENIA and CRAFT, respectively, resulting in the highest accuracies on both experimental corpora. Note that for PTB, CNN-based character-level word embeddings \cite{ma-hovy:2016:P16-1} only provided a 0.1\% improvement to  BiLSTM-CRF \cite{HuangXY15}.  
 The larger improvements on GENIA and CRAFT show that character-level word embeddings are specifically   useful to capture rare or unseen words in biomedical text data. 
 Character-level word embeddings  are useful for  morphologically rich languages \cite{plank-sogaard-goldberg:2016:P16-2,nguyen-dras-johnson-2017}, and although English is not morphologically rich, the biomedical domain contains a wide variety of morphological variants of domain-specific terminology \cite{Liu2012}. Words  tagged  incorrectly are largely associated with gold tags \textit{NN}, \textit{JJ} and \textit{NNS}; many are abbreviations which are also out-of-vocabulary. \CHANGEA{It is typically difficult  for   character-level  word embeddings to capture those unseen abbreviated words \cite{Maryam:2017}.}

On both GENIA and CRAFT, BiLSTM-CRF with character-level word embeddings obtains the highest accuracy scores. These are just  0.1\% absolute higher than the accuracies of NLP4J-POS.  
Note that small variations in POS tagging performance are not a critical factor in parsing performance \cite{seddah-EtAl:2010:SPMRL}. 
In addition, we find that NLP4J-POS obtains 30-time  faster training and testing speed. Hence for the dependency parsing task, we use NLP4J-POS to  perform 20-way jackknifing \cite{koo-carreras-collins:2008:ACLMain}  to generate POS tags on training data and to predict POS tags on development and test sets.

\begin{figure*}[!t]
\centering
$\begin{array}{ll}
\includegraphics[height=5cm,width=8cm]{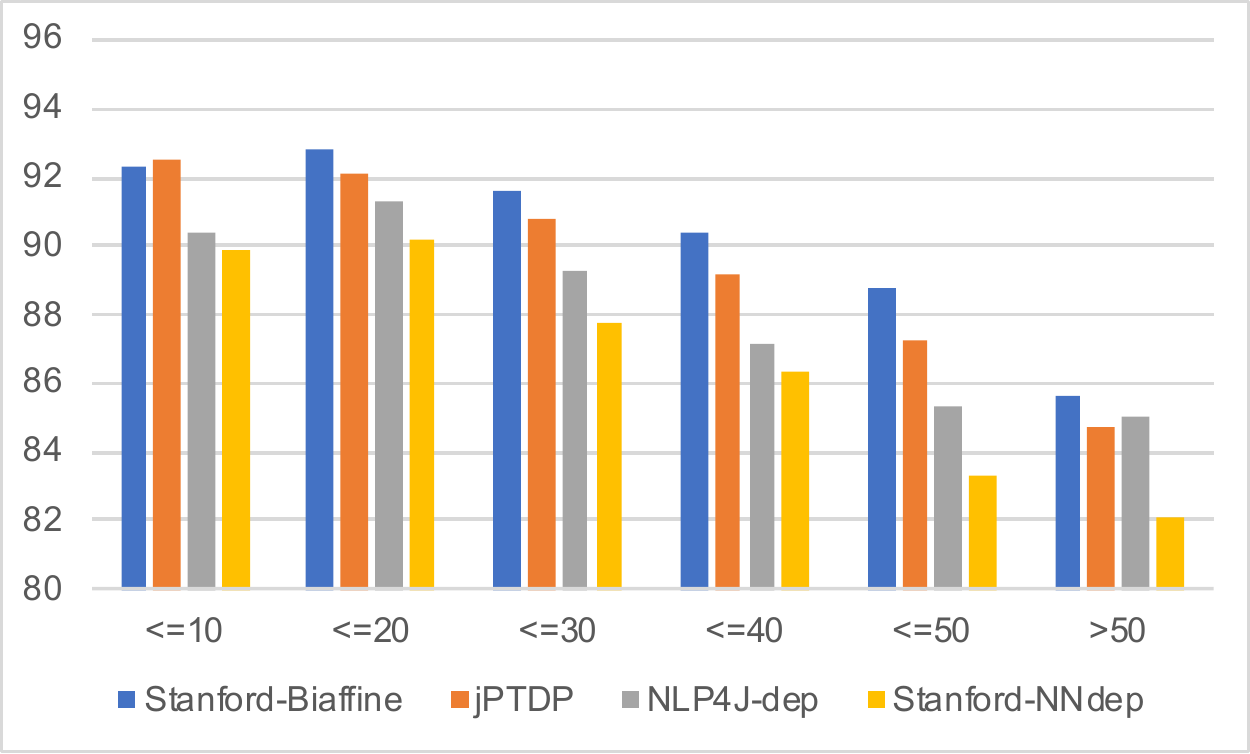} &
\includegraphics[height=5cm,width=8cm]{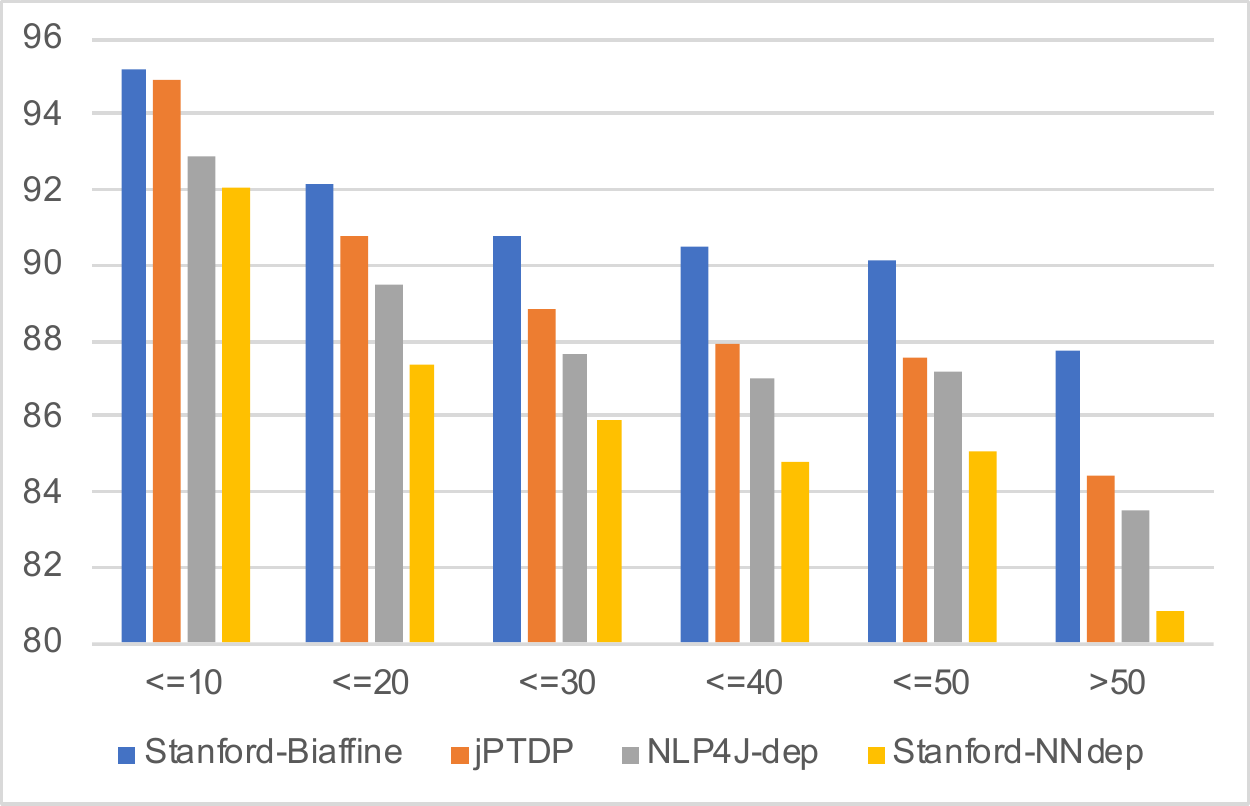} 
\\
\mbox{({GENIA -- sentence length})} & \mbox{({CRAFT -- sentence length})}
\end{array}$
\caption{LAS scores by sentence length.}
\label{fig:LASbyLength}
\end{figure*}

\begin{figure*}[!t]
\centering
$\begin{array}{ll}
\includegraphics[height=5cm,width=8cm]{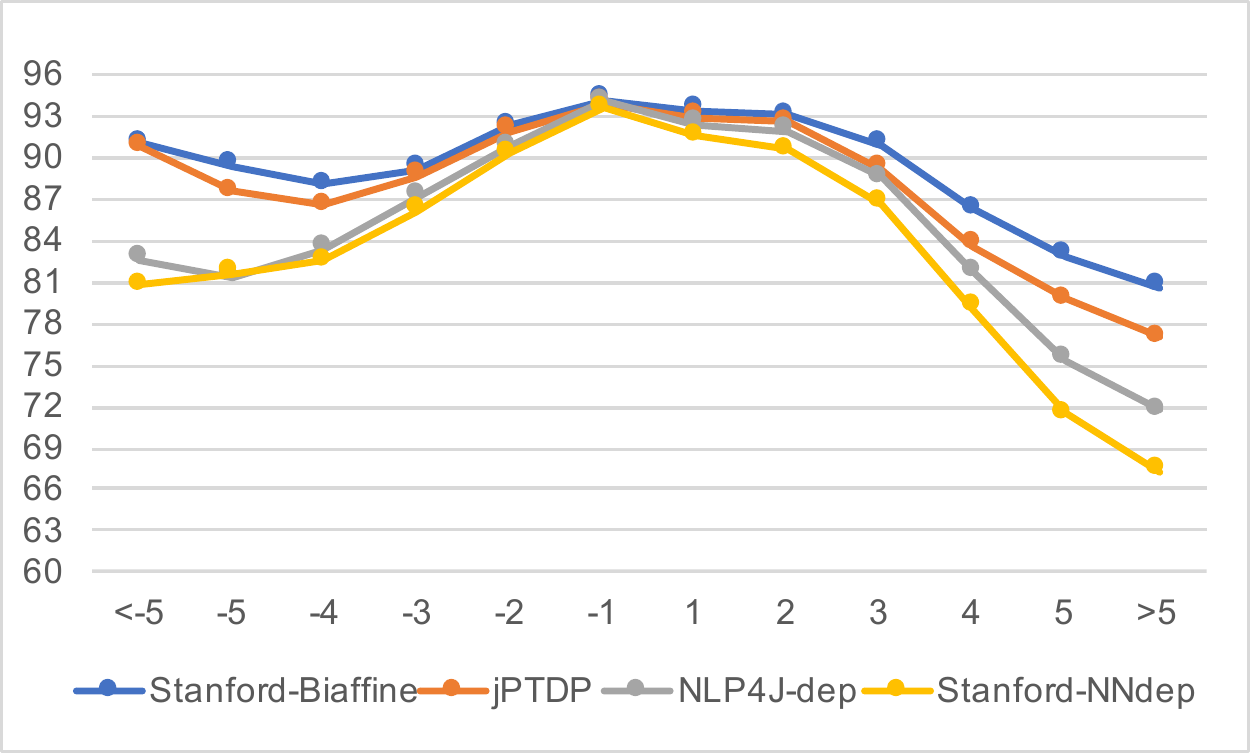} &
\includegraphics[height=5cm,width=8cm]{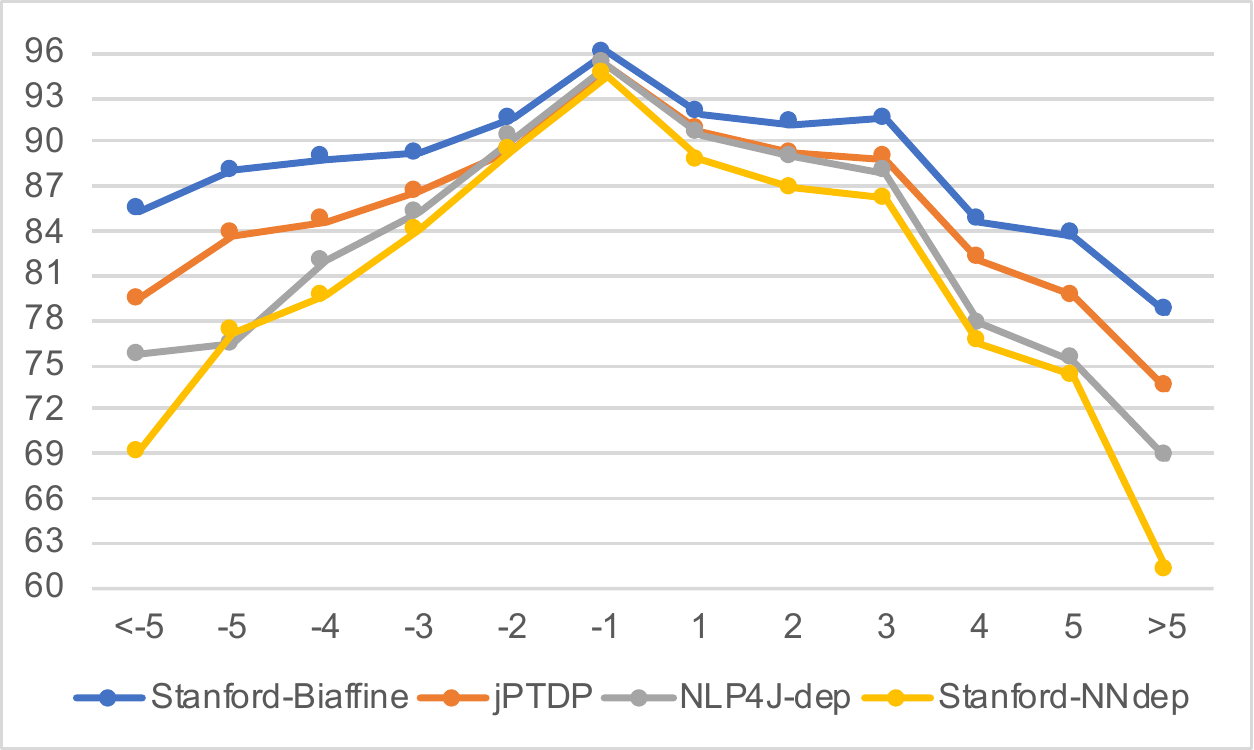} 
\\
\mbox{({GENIA -- dependency distance})} & \mbox{({CRAFT -- dependency distance})}
\end{array}$
\caption{LAS (F1) scores by dependency distance.}
\label{fig:LASbyDistance}
\end{figure*}

\subsection*{Overall dependency parsing results}

We present the LAS and UAS scores of different parsing models in Table \ref{tab:mainresults}. 
{The first five rows show parsing results on the GENIA test set of ``pre-trained'' parsers. 
The first two rows  present scores of the pre-trained   Stanford NNdep and Biaffine v1  models  with POS tags predicted by the pre-trained Stanford tagger \cite{Toutanova:2003:FPT:1073445.1073478}, while  the next two rows 3-4 present scores of    these pre-trained models     with POS tags predicted by NLP4J-POS.} Both pre-trained NNdep and Biaffine models were trained on a dependency treebank of 40K sentences, which was 
 converted from the English PTB sections 2--21. 
{The fifth row shows  scores of BLLIP+Bio, the BLLIP reranking constituent parser \cite{charniak-johnson:2005:ACL} with an improved self-trained biomedical parsing model \cite{david2010}.}  
We use the Stanford conversion toolkit (v3.5.1) to generate dependency trees with the basic Stanford dependencies  
and use the data split on GENIA as  used in \cite{david2010}, therefore  parsing scores are comparable.  
The remaining rows show results  of our retrained dependency parsing models. 

On GENIA, among pre-trained models, BLLIP obtains highest results. This model, unlike the other pre-trained models, was trained using GENIA, so this result is unsurprising. The pre-trained Stanford-Biaffine  (v1) model produces lower scores than the pre-trained Stanford-NNdep model on GENIA. It is also unsurprising because the pre-trained Stanford-Biaffine utilizes pre-trained  word vectors  which were learned from newswire  corpora.  
Note that the pre-trained NNdep and Biaffine models result in no significant performance differences irrespective of the source of POS tags (i.e.\ the pre-trained Stanford tagger at 98.37\% vs. the retrained NLP4J-POS model at 98.80\%).

Regarding the retrained  parsing models, on both GENIA and CRAFT,   Stanford-Biaffine  achieves the highest parsing results with LAS at 91.23\% and UAS at 92.64\%  on GENIA, and LAS at 90.77\% and UAS at 92.67\% on CRAFT, computed without punctuations.  
Stanford-NNdep  obtains the lowest  scores; about 3.5\% and 5\% absolute lower  than Stanford-Biaffine on GENIA and CRAFT, respectively.  jPTDP is ranked second, obtaining about   1\% and 2\% lower scores than Stanford-Biaffine  and 1.5\% and 1\% higher scores  (without punctuation) than   NLP4J-dep on GENIA and CRAFT, respectively.   \CHANGEA{Table \ref{tab:mainresults} also shows that the best  parsing model  Stanford-Biaffine obtains    about 1\% absolute improvement when using gold POS tags instead of predicted POS tags. }

\section*{Parsing result analysis}\label{sssec:accuracyana}
Here we present a detailed  analysis of the parsing results obtained by the retrained  models  \CHANGEA{with predicted POS tags.} 
For simplicity,  the following more detailed analyses report LAS scores, computed  without punctuation. Using UAS  scores or computing with punctuation does not reveal any additional information.

\subsection*{Sentence length}

Figure \ref{fig:LASbyLength} presents LAS scores by sentence length in bins of length 10.  As expected, all parsers produce better results for shorter sentences on both corpora; longer sentences are likely to have longer dependencies which are typically harder to predict precisely. 
 Scores drop by about 10\% for sentences longer than 50 words, relative to short sentences ${<=}10$ words.  Exceptionally, on GENIA we find lower scores for the shortest sentences than for the  
sentences from 11 to 20 words. This is probably because abstracts tend not to contain short sentences: (i) as shown in Table \ref{tab:statistics},  
the proportion of sentences in the first bin is very low at 3.5\% on GENIA (cf.\ 17.8\% on CRAFT), and (ii)   sentences in the  first  bin on GENIA are relatively long, with an average length of 9 words (cf.\ 5 words in CRAFT).

\subsection*{Dependency distance}

Figure \ref{fig:LASbyDistance}  shows LAS (F1) scores corresponding to  the dependency distance $i - j$, between a dependent $w_i$ and its head $w_j$, where $i$ and $j$ are consecutive indices of words in a sentence.  Short dependencies are often modifiers of nouns such
as determiners or adjectives or pronouns modifying their direct neighbors, while longer  dependencies typically represent modifiers of the root or the main verb   \cite{mcdonald-nivre:2007:EMNLP-CoNLL2007}.
All parsers obtain higher scores for left dependencies than  for  right dependencies. This is not completely  unexpected as English is strongly head-initial. 
In addition, the gaps between LSTM-based models (i.e.\ Stanford-Biaffine and jPTDP) and non-LSTM models (i.e.\ NLP4J-dep and Stanford-NNdep) are larger for the long dependencies than for the shorter ones, as LSTM architectures can preserve long range information \cite{Graves2008}.

  \begin{table}[!t]
  \caption{\CHANGEA{LAS (F1) scores  of Stanford-Biaffine on GENIA,  by frequent dependency labels in the  left dependencies. ``Prop.'' denotes the occurrence proportion in each distance bin.}}
  \centering
\resizebox{8cm}{!}{
\def\arraystretch{1.05}
\begin{tabular}{l|l|l|l|l|l|l}
\hline
\multirow{2}{*}{Type} & \multicolumn{2}{c|}{\bf ${<}$ $-5$} & \multicolumn{2}{c|}{\bf $-5$} & \multicolumn{2}{c}{\bf $-4$} \\
\cline{2-7}
& Prop.& LAS& Prop.& LAS& Prop.& LAS \\
\hline
advmod & 7.2 & \textbf{94.62} & 4.2 & 90.91 & 4.6 & 88.52     \\
amod & 4.8 & 74.19 & 8.1 & 80.00 & 17.5 & \textbf{86.09}   \\
det & 4.3 & 85.71 & 17.7 & \textbf{91.43} & 21.3 & 88.97   \\
mark & 15.4 & 98.49 & 11.5 & \textbf{98.90} & 6.4 & 97.62     \\
nn & 4.7 & 74.38 & 15.7 & \textbf{77.42} & 16.6 & 76.71    \\
nsubj & 28.2 & 93.96 & 19.0 & 94.67 & 15.3 & \textbf{96.52}    \\
nsubjpass & 15.9 & \textbf{95.38}  & 11.3 & 92.13 & 3.9 & 86.27     \\
prep & 11.9 & 96.10 & 6.7 & \textbf{98.11} & 2.6 & 88.24    \\
\hline 
\end{tabular}
}
\label{tab:depbylasposjPTDP}
\end{table}

On both corpora,  higher scores are also associated with shorter distances. There is one  surprising  exception:  on GENIA, in  
distance bins of $-4$, $-5$ and ${<}$ $-5$, Stanford-Biaffine and jPTDP obtain higher scores for longer distances.   This may result from the structural characteristics of sentences in the GENIA corpus. 
 \CHANGEA{Table \ref{tab:depbylasposjPTDP} details the scores  of Stanford-Biaffine in terms of the most frequent dependency labels in these left-most dependency bins.}  We find \textit{amod} and \textit{nn} are the two most difficult to predict dependency relations \CHANGEA{(the same finding applied to jPTDP)}. They appear much more frequently in the bins $-4$ and $-5$ than in  bin ${<}$ $-5$, explaining the higher overall score for bin ${<}$ $-5$. 

\begin{table}[!t]
\caption{LAS  by  the basic Stanford dependency labels on GENIA. ``Avg.'' denotes the averaged score of the four dependency parsers. }
\centering
\begin{tabular}{l|c|c|c|c|c}
\hline
Type & Biaffine	& jPTDP	& NLP4J	& NNdep & Avg. \\
\hline
advmod	 & \textbf{87.38}	 & 86.77	 & 87.26	 & 83.86 & 86.32 \\
amod	 & \textbf{92.41}	 & 92.21	 & 90.59	 & 90.94 & 91.54 \\
appos & 	\textbf{84.28} & 	83.25 & 	80.41 & 	77.32 &81.32  \\
aux & 	98.74	 & \textbf{99.28} & 	98.92 & 	97.66 & 98.65 \\
auxpass & 	99.32	 & 99.32 & 	\textbf{99.49}	 & 99.32 & 99.36 \\
cc & 	\textbf{89.90}	 & 86.38 & 	82.21	 & 79.33 &84.46  \\
conj & 	\textbf{83.82}	 & 78.64	 & 73.32	 & 69.40 & 76.30 \\
dep	 & 40.49 & 	\textbf{41.72}	 & 40.04	 & 31.66 &  38.48\\
det	 & \textbf{97.16}	 & 96.68	 & 95.46 & 	95.54  &96.21 \\
dobj & 	\textbf{96.49}	 & 95.87	 & 94.90	 & 92.18  &94.86 \\
mark & \textbf{94.68}  & 90.38 & 89.62 & 90.89  & 91.39 \\
nn	 & 90.07	 & \textbf{90.25}	 & 88.22 & 	88.97 & 89.38 \\
nsubj	 & \textbf{95.83}	 & 94.71	 & 93.18	 & 90.75 & 93.62 \\
nsubjpass & 	\textbf{95.56}	 & \textbf{95.56}	 & 92.05 & 	90.94 &  93.53\\
num	 & 89.14	 & 85.97	 & {90.05} & 	\textbf{90.27} &  88.86\\
pobj	 & \textbf{97.04} & 	96.54 & 	96.54	 & 95.13 & 96.31 \\
prep & 	\textbf{90.54} & 	89.93 & 	89.18 & 	88.31 &  89.49\\
root	 & \textbf{97.28} & 	97.13 & 	94.78 & 	92.87 & 95.52 \\
\hline
\end{tabular}
\label{tab:LAS_GENIA_DepTypes}
\end{table}

\begin{table}[!t]
\caption{LAS  by the CoNLL 2008 dependency labels on CRAFT. }
\centering
\begin{tabular}{l|c|c|c|c|c }
\hline
Type & Biaffine	& jPTDP	& NLP4J	& NNdep & Avg. \\
\hline
ADV		 & \textbf{79.20}		 & 77.53	 & 	75.58	 & 	71.64 & 75.99 \\
AMOD	 & 	\textbf{86.43}	 & 	83.45	 & 	85.00	 & 	82.98 &84.47  \\
CONJ		 & \textbf{91.73}		 & 88.69		 & 85.42		 & 83.34 &  87.30\\
COORD		 & \textbf{88.47}	 & 	84.75	 & 	79.42		 & 76.38  &82.26 \\
DEP		 & \textbf{73.23}		 & 67.96		 & 62.83		 & 52.43 & 64.11 \\
LOC		 & \textbf{70.70}		 & 68.91	 & 	68.64		 & 61.35 & 67.40 \\
NMOD		 & \textbf{92.55}	 & 	91.19	 & 	90.77	 & 	90.04 &  91.14\\
OBJ		 & \textbf{96.51}	 & 	94.53		 & 93.85		 & 91.34 &  94.06\\
PMOD		 & \textbf{96.30}		 & 94.85	 & 	94.52		 & 93.44 &94.78  \\
PRD		 & \textbf{93.96}	 & 	90.11		 & 92.49		 & 90.66 & 91.81 \\
PRN	 & 	\textbf{62.11	}	 & 61.30		 & 49.26	 & 	46.96 &  54.91\\
ROOT		 & \textbf{98.15}		 & 97.20		 & 95.24		 & 91.27 & 95.47 \\
SBJ		 & \textbf{95.87}		 & 93.03	 & 	91.82	 & 	90.11 & 92.71 \\
SUB		 & \textbf{95.18}	 & 	91.81	 & 	91.81	 & 	89.64 &  92.11\\
TMP		 & \textbf{78.76}	 & 	68.81	 & 	65.71		 & 59.73 &68.25 \\
VC		 & \textbf{98.84}		 & 97.50		 & 98.09		 & 96.09 & 97.63 \\
\hline
\end{tabular}
\label{tab:LAS_CRAFT_DepTypes}
\end{table}

\subsection*{Dependency label}

Tables \ref{tab:LAS_GENIA_DepTypes} and \ref{tab:LAS_CRAFT_DepTypes} present LAS scores for the most frequent dependency relation types on GENIA and CRAFT, respectively. In most cases, Stanford-Biaffine obtains the highest score for each relation type on both corpora with the following exceptions: on GENIA, jPTDP gets the highest results to \textit{aux}, \textit{dep} and \textit{nn} (as well as \textit{nsubjpass}), while NLP4J-dep and NNdep obtain the highest scores for \textit{auxpass} and \textit{num}, respectively. 
 On GENIA the labels associated with the highest average LAS scores (generally ${>}90\%$) are \textit{amod}, \textit{aux}, \textit{auxpass}, \textit{det}, \textit{dobj}, \textit{mark},  \textit{nsubj}, \textit{nsubjpass}, \textit{pobj} and \textit{root} whereas on CRAFT they are \textit{NMOD}, \textit{OBJ}, \textit{PMOD}, \textit{PRD}, \textit{ROOT}, \textit{SBJ}, \textit{SUB} and \textit{VC}. These labels either correspond to short dependencies (e.g.\ \textit{aux}, \textit{auxpass} and \textit{VC}), have strong lexical indications (e.g.\ \textit{det}, \textit{pobj} and \textit{PMOD}), or  occur very often (e.g.\ \textit{amod}, \textit{subj}, \textit{NMOD} and \textit{SBJ}). 

Those relation types with the lowest LAS scores (generally ${<}70\%$)  are \textit{dep} on GENIA and \textit{DEP}, \textit{LOC}, \textit{PRN} and \textit{TMP} on CRAFT;  \textit{dep}/\textit{DEP} are very general labels while \textit{LOC}, \textit{PRN} and \textit{TMP}  are among the least frequent labels. Those types also associate to the biggest variation of obtained accuracy across parsers (${>}8\%$). In addition, the coordination-related labels \textit{cc}, \textit{conj}/\textit{CONJ} and \textit{COORD} show large variation across parsers. These  9  mentioned relation labels  generally correspond to long dependencies. 
Therefore, it is not surprising that BiLSTM-based models Stanford-Biaffine and jPTDP can produce much higher accuracies on these labels than non-LSTM models NLP4J-dep and NNdep.

The remaining types are either relatively rare labels (e.g.\ \textit{appos}, \textit{num} and \textit{AMOD}) or more frequent labels  but with a varied distribution of dependency distances (e.g.\ \textit{advmod}, \textit{nn}, and \textit{ADV}).

\begin{table*}[!t]
\centering
\caption{LAS by POS tag of the dependent.}
\def\arraystretch{1.05}
\begin{tabular}{l|c|c|c|c|c|c|c|c}
\hline
\multirow{2}{*}{Type} & \multicolumn{4}{c|}{\bf GENIA}  & \multicolumn{4}{c}{\bf CRAFT} \\
\cline{2-9}
& Biaffine	& jPTDP	& NLP4J	& NNdep & Biaffine	& jPTDP	& NLP4J	& NNdep\\
\hline
CC	 & \textbf{89.71}	 & 86.70	 & 82.75	 & 80.20	 & \textbf{89.01}	 & 85.45	 & 79.99	 & 77.45 \\
CD	 & \textbf{81.83}	 & 79.30	 & 79.78	 & 79.30	 & \textbf{88.03} & 	85.17 & 	84.22	 & 79.77 \\
DT	 & \textbf{95.31}	 & 95.09	 & 93.99	 & 93.08	 & \textbf{98.27}	 & 97.39	 & 97.18	 & 96.77 \\
IN	 & \textbf{90.57}	 & 89.50	 & 88.41 & 	87.58	 & \textbf{81.79}	 & 79.32	 & 78.43	 & 75.97 \\
JJ	 & \textbf{90.17} & 	89.35 & 	88.30	 & 87.76	 & \textbf{94.24} & 	92.91 & 	92.50 & 	91.70 \\
NN	 & \textbf{90.69}	 & 89.92 & 	88.26 & 	87.62 & 	\textbf{91.24} & 	89.28	 & 88.32 & 	87.48 \\
NNS	 & \textbf{93.31}	 & 92.32 & 	91.33 & 	87.91	 & \textbf{95.07}	 & 92.57	 & 90.91 & 	88.30 \\
RB	 & \textbf{88.31} & 	86.92 & 	87.73 & 	84.61 & 	\textbf{84.41} & 	81.98 & 	82.13	 & 76.99 \\
TO	 & 90.97	 & {91.50}	 & \textbf{92.04} & 88.14 & 	90.16 & 	85.83 & 	\textbf{90.55} & 	83.86 \\
VB	 & \textbf{89.68} & 	87.84 & 	85.09	 & 83.49 & 	\textbf{98.86} & 	\textbf{98.86} & 	98.67	 & 96.38 \\
VBD	 & \textbf{94.60} & 93.85 & 	90.97	 & 90.34	 & \textbf{94.74} & 	93.21 & 	90.03	 & 86.86 \\
VBG	 & \textbf{82.67} & 	79.47 & 	79.20	 & 72.27	 & \textbf{85.51}	 & 81.33 & 	81.15 & 	75.57 \\
VBN	 & \textbf{91.42} & 	90.53 & 	88.02 & 	85.51	 & \textbf{93.22}	 & 91.24	 & 90.25 & 	88.04 \\
VBP	 & \textbf{94.46}	 & 93.88	 & 92.54	 & 90.63	 & \textbf{93.54} & 	91.18	 & 88.98	 & 84.09 \\
VBZ	 & \textbf{96.39} & 	94.83 & 	93.57 & 	92.48 & 	\textbf{93.42} & 	88.77 & 	87.67 & 	84.25 \\
\hline
\end{tabular}
\label{tab:laspos}
\end{table*}

\subsection*{POS tag of the dependent}

Table \ref{tab:laspos} analyzes the LAS scores by the most frequent POS tags (across two corpora) of the dependent. Stanford-Biaffine achieves the highest scores on all these tags except \textit{TO} where the traditional feature-based model NLP4J-dep obtains the highest score (\textit{TO} is relatively rare tag in GENIA and is the least frequent tag in CRAFT among  tags listed in Table \ref{tab:laspos}). Among listed tags \textit{VBG} is the least and second least frequent one in GENIA and CRAFT, respectively, and generally associates to longer dependency distances. So, it is reasonable that the lowest scores we obtain on both corpora are accounted for by \textit{VBG}. The coordinating conjunction tag \textit{CC} also often corresponds to long  dependencies,  thus resulting in biggest ranges across parsers on both GENIA and CRAFT. The results for \textit{CC}  are
consistent with the results obtained for the dependency labels  \textit{cc} in 
 Table \ref{tab:LAS_GENIA_DepTypes} and   \textit{COORD}  in Table  \ref{tab:LAS_CRAFT_DepTypes} because they are coupled to each other.

On the remaining POS tags, we generally find similar patterns across parsers and corpora, except for  \textit{IN} and \textit{VB} where  parsers produce 8+\% higher scores for \textit{IN} on GENIA than on CRAFT, and vice versa producing 9+\% lower scores for \textit{VB} on GENIA.  
This is because on GENIA,   \textit{IN} is  mostly coupled with the dependency label \textit{prep} at a rate  of 90\%  (thus their corresponding LAS  scores in tables \ref{tab:laspos}   and \ref{tab:LAS_GENIA_DepTypes} are consistent), while on CRAFT \textit{IN} is  coupled to a more varied distribution of dependency labels such as \textit{ADV} with a  rate at 20\%, \textit{LOC} at 14\%, \textit{NMOD} at 40\% and \textit{TMP} at 5\%.  Regarding \textit{VB},  on CRAFT it usually associates to a short dependency distance of 1 word (i.e.\ head and dependent words are next to each other) with a rate at 80\%, and to a distance of 2 words at  15\%, while on GENIA it associates with longer dependency distances with a rate at 17\% for the distance of 1 word, 31\% for the distance of 2 words and 34\% for a distance of $>5$ words. So, parsers obtain much higher scores for \textit{VB} on CRAFT than on GENIA.

\begin{table}[!t]
\centering
\caption{Error examples. ``H.'' denotes the  head index of the current word.  }
\resizebox{8cm}{!}{
\def\arraystretch{1.05}
\setlength{\tabcolsep}{0.4em}
\begin{tabular}{ll|lll|lll}
\hline
\multirow{2}{*}{\bf ID}& \multirow{2}{*}{\bf Form}&\multicolumn{3}{c|}{\bf Gold} & \multicolumn{3}{c}{\bf Prediction}\\
\cline{3-8}
 &   & POS & H. & DEP & POS & H. & DEP \\
\hline
19	 &	both	 &		CC	 &		24 	 &	preconj	 & CC	&	\textbf{21}	& preconj	\\
20	 &	the &	DT	 &		24 &	det & DT	&	\textbf{21} &	\textbf{dep}	\\
21	 &	POU(S)	 &		JJ &		24  &		amod & \textbf{NN}	& \textbf{18}	& \textbf{pobj}  \\
22 &		and &		CC	 &	21	 &	cc	  &  CC	&	21	& cc\\
23  &		POU(H)	 &		NN	 &		21	 &	conj	 &	 NN	&	21 &	conj \\
24	 &	domains &		NNS &	18 &	pobj 	 &	 NNS	 &	\textbf{21}	 & \textbf{dep}	\\
\hline
23	 & the	& DT	 & 26 & 	det	& DT	& 	\textbf{27}	 & det \\
24	& Oct-1-responsive	& 	JJ & 	26 & 	amod	& JJ	& 	\textbf{27} &	amod\\
25	 & octamer	& 	NN	& 	26	& nn & NN &\textbf{27}	& nn \\
26	& sequence	& NN	& 22	& pobj	& NN	&	\textbf{27} & 	\textbf{nn} \\ 
27	 & ATGCAAAT	& 	NN	& 	26 & 	dep & NN	&	\textbf{22} &	\textbf{pobj}	\\
\hline
\end{tabular}
}
\label{tab:error}
\end{table}

\subsection*{Error analysis}
We analyze \CHANGEA{token-level parsing errors that occur consistently across all parsers (i.e.\ the intersection set of errors)}, and find that there are few common error patterns.  \CHANGEA{The first one is related to incorrect POS tag prediction (8\% of the intersected parsing errors on GENIA and 12\% on CRAFT are coupled with incorrect predicted POS tags).} For example,  the word token ``domains'' is the head of the phrase ``both the POU(S) and POU(H) domains''   \CHANGEA{in Table  \ref{tab:error}.} We also have two OOV word tokens ``POU(S)'' and ``POU(H)'' which abbreviate ``POU-specific'' and ``POU homeodomain'', respectively.  NLP4J-POS (as well as all other POS taggers) produced an incorrect tag of NN rather than adjective (JJ) for ``POU(S)''. As ``POU(S)'' is predicted to be a noun, all parsers make an incorrect  prediction that it is the phrasal head,  \CHANGEA{thus also resulting in errors to remaining dependent words in the phrase.}

The second error type occurs on noun phrases such as ``the Oct-1-responsive octamer sequence ATGCAAAT''  \CHANGEA{(in Table  \ref{tab:error})} and ``the herpes simplex virus Oct-1 coregulator VP16'', \CHANGEA{commonly referred to as appositive structures,} where the second to last noun (i.e.\ ``sequence'' and ``coregulator'') is considered to be the phrasal head, rather than the last noun. However, such phrases are relatively rare and all parsers predict the last noun as the head.  

\CHANGEA{The third error type is related to the relation labels   \textit{dep/DEP}. We manually re-annotate every case where all parsers agree on the dependency label for a dependency arc with the same dependency label, where this label disagrees with the gold label \textit{dep/DEP}
(these cases are about 3.5\% of the parsing errors intersected across all parsers on GENIA  and 0.5\% on CRAFT). Based on this manual review, we find that about 80\% of these cases appear to be labelled correctly, despite not agreeing with the gold standard. In other words, the gold standard appears to be in error in these cases.
This result is not completely unexpected because when converting from constituent treebanks to dependency treebanks, the general dependency label \textit{dep/DEP} is usually assigned due to  limitations in the automatic conversion toolkit.}

\section*{Parser comparison on event extraction}
We present an extrinsic evaluation of the four dependency parsers for the downstream task of biomedical event extraction.
\subsection*{Evaluation setup}
Previously, Miwa et al.\ \cite{miwa2010} adopted the BioNLP 2009 shared task on biomedical event extraction \cite{BioNLP2009} to compare the task-oriented performance of six ``pre-trained'' parsers with  3 different types of dependency representations. 
However, their evaluation setup requires use of  
a currently unavailable event extraction system.  
Fortunately, the extrinsic parser evaluation (EPE 2017)  shared task  aimed to evaluate different dependency representations by comparing their performance on downstream tasks \cite{epe2017}, including a biomedical event extraction  task   \cite{epe2017bio}.  
We thus follow the experimental setup used there;  employing the Turku Event Extraction System (TEES, \cite{bjorne-EtAl:2009:BioNLP-ST}) to assess the impact of parser differences on biomedical relation extraction.\footnote{{\url{https://github.com/jbjorne/TEES/wiki/EPE-2017}}}

EPE 2017  uses the BioNLP 2009  shared task dataset \cite{BioNLP2009}, which was derived   from  the GENIA treebank corpus (800, 150 and 260 abstract files used for BioNLP 2009 training, development and test, respectively).\footnote{678 of 800 training,  132 of  150 development and 248 of 260 test files are included in the GENIA treebank training set.}  
We only need to provide dependency parses of raw texts using the pre-processed tokenized and sentence-segmented data provided by the EPE 2017 shared task. For the Stanford-Biaffine, NLP4J-dep and Stanford-NNdep parsers that require predicted POS tags, we  use the retrained NLP4J-POS model  to generate POS tags.  We then produce parses using retrained dependency parsing models.

TEES is then trained for the BioNLP 2009 Task 1 using the training data, and is  evaluated on the development data (gold event annotations are only available to public for training and development sets). To obtain test set performance, we use an  online evaluation system. 
The online evaluation system for the BioNLP 2009 shared task is currently not available. Therefore, we employ the online evaluation system for the BioNLP 2011 shared task \cite{kim-EtAl:2011:BioNLP-ST1}   with the ``abstracts only'' option.\footnote{\url{http://bionlp-st.dbcls.jp/GE/2011/eval-test/eval.cgi}}  \CHANGEA{The score is reported using the approximate span \& recursive evaluation strategy \cite{BioNLP2009}.}

 \begin{table}[!t]
     \caption{UAS and LAS (F1) scores of re-trained models on the pre-segmented BioNLP-2009 development sentences  which contain event interactions. Scores  are computed on all tokens using the evaluation script from the CoNLL 2017 shared task   \protect{\cite{zeman-EtAl:2017:K17-3}}.} 
    \centering
    \begin{tabular}{l|llll}
    \hline
       Metric  & Biaffine  & jPTDP  & NLP4J  & NNdep \\
       \hline 
        UAS & 95.51 & 93.14 & 92.50 & 91.02 \\
        LAS & 94.82 & 92.18 & 91.96 & 90.30 \\
        \hline
    \end{tabular}

    \label{tab:devscores}
\end{table}

 \begin{table*}[!t]
\centering
\caption{Biomedical event extraction results. The subscripts denote  results for which TEES is trained without the dependency labels. }
\def\arraystretch{1.05}
\begin{tabular}{ll|lll|lll}
\hline
\multicolumn{2}{c|}{\multirow{2}{*}{\bf System}} & \multicolumn{3}{c|}{\bf Development} & \multicolumn{3}{c}{\bf Test}\\
\cline{3-8}
& &   R & P & F$_1$  &   R & P & F$_1$\\
\hline
& Stanford\&Paris & 49.92 & 55.75 &  52.67 & 45.03 & 56.93  &  50.29 \\
& BLLIP+Bio & 47.90  &   61.54  &  53.87\textsubscript{52.35} & 41.45    & 60.45    & 49.18\textsubscript{49.19} \\
\hline
\multirow{4}{*}{\rotatebox[origin=c]{90}{GENIA}} &  Stanford-Biaffine-v2  & 50.53 &    56.47    & 53.34\textsubscript{\textbf{53.18}} & 43.87   &  56.36  &  49.34\textsubscript{\textbf{49.47}}  \\
 & jPTDP-v1  & 49.30  &   58.58  &  \textbf{53.54}\textsubscript{52.08} & 42.11  &  54.94  &  47.68\textsubscript{48.88} \\
 & NLP4J-dep 	 &    51.93  &  55.15   &  53.49\textsubscript{52.20} & 45.88  &  55.53  &  \textbf{50.25}\textsubscript{49.08} \\
 & Stanford-NNdep  &  46.79  &   60.36  &   52.71\textsubscript{51.38} &  40.16  &  59.75  &  48.04\textsubscript{48.51}\\
\hline
\multirow{4}{*}{\rotatebox[origin=c]{90}{CRAFT}} &  Stanford-Biaffine-v2 & 49.47  &  57.98 &     53.39\textsubscript{\textbf{52.98}} &  42.08 &   58.65   &  \textbf{49.00}\textsubscript{\textbf{49.84}}   \\
 & jPTDP-v1  &  49.36  &  58.22  &  \textbf{53.42}\textsubscript{52.01} & 40.82  &  58.57  &  48.11\textsubscript{49.57} \\
 & NLP4J-dep 	 &  48.91  &  53.13   &  50.93\textsubscript{51.03} & 41.95  &  51.88 &   46.39\textsubscript{47.46}  \\
 & Stanford-NNdep  & 46.34   &  56.83  &  51.05\textsubscript{51.01} & 38.87  &  59.64  &  47.07\textsubscript{46.38} \\
\hline
\end{tabular}
\label{tab:bionlp}
\end{table*}

\subsection*{Impact of parsing on event extraction}

 Table \ref{tab:devscores} presents the intrinsic UAS and LAS (F1) scores on the  pre-processed  segmented BioNLP 2009 development sentences (i.e.\ scores with respect to  predicted segmentation), for which these sentences contain event interactions.  These scores are higher than those presented in Table \ref{tab:mainresults} because most part of the BioNLP 2009 dataset is extracted from  the GENIA treebank training set. 
 Although gold event annotations  in the BioNLP 2009 test set are not available to public, it is likely that we would obtain the similar intrinsic UAS and LAS scores on the pre-processed  segmented test  sentences containing event interactions.

Table \ref{tab:bionlp} compares parsers with respect to the EPE 2017 biomedical event extraction task   \cite{epe2017bio}. The first row presents the score of the Stanford\&Paris  team  \cite{stanfordparis2017}; the highest official score obtained on the test set.  
Their system  
used the  Stanford-Biaffine parser (v2) trained on a dataset  combining  
PTB, Brown corpus, and GENIA treebank data.\footnote{The EPE 2017  shared task \cite{epe2017} focused on evaluating different dependency representations in downstream tasks, not on comparing different parsers. Therefore each participating team  employed only one parser, either a dependency graph or tree parser.  Only the  Stanford\&Paris team \cite{stanfordparis2017} employ GENIA data, obtaining the highest biomedical event extraction score.}  
 The second row presents our score for the pre-trained BLLIP+Bio model;  
  remaining rows show scores using re-trained parsing models.

The results for parsers trained with the GENIA treebank (Rows 1-6, Table \ref{tab:bionlp}) are generally higher than for parsers trained on CRAFT. This is logical because the BioNLP 2009 shared task dataset was a subset of  the GENIA corpus.
However, we find that the differences in intrinsic parsing results  as presented in tables \ref{tab:mainresults} and \ref{tab:devscores} do not consistently explain the differences in extrinsic biomedical event extraction performance, extending preliminary related observations in prior work \cite{nguyen-verspoor:2018:K18-2,mackinlay2013extracting}. Among the four dependency parsers trained on GENIA,  Stanford-Biaffine, jPTDP and NLP4J-dep produce similar event extraction scores on the  development set, while  on the the test set jPTDP and NLP4J-dep obtain  the lowest  and  highest scores, respectively.

Table  \ref{tab:bionlp}  also summarizes the results with the dependency structures only (i.e.\ results without dependency relation labels; replacing all predicted dependency labels by  ``UNK''  before training TEES). In most cases, compared to using dependency labels, event extraction scores drop on the development set (except NLP4J-dep trained on CRAFT), while they increase on the test set (except NLP4J-dep trained on GENIA and Stanford-NNdep trained on CRAFT). Without dependency labels,  better event extraction scores on the development set corresponds to better scores on the test set. In addition, the differences in these event extraction scores without dependency labels are more consistent with the parsing performance differences than the scores with dependency labels. 

These findings show that variations in dependency  representations strongly affect event extraction performance.  Some (predicted) dependency labels are likely to be particularly useful for extracting events, while others hurt performance. Also, investigating $\sim$20 frequent dependency labels in each dataset as well as some possible combinations between them could lead to an enormous number of additional experiments. We believe a detailed analysis of the interaction between those  labels in a downstream application task deserves another research paper with a more careful analysis. Here, one contribution of our paper could be seen to be that we highlight the need for further research  in this direction.
 
\section*{Conclusion}

We have presented a detailed empirical study comparing SOTA traditional feature-based and neural network-based models for POS tagging and dependency parsing in the biomedical context.  
In general, the neural models outperform the feature-based models on two benchmark biomedical corpora GENIA and CRAFT. In particular,  BiLSTM-CRF-based models with character-level word embeddings produce highest POS tagging accuracies  which are slightly better than NLP4J-POS, while the   Stanford-Biaffine parsing model obtains significantly better  result than other parsing models.    

We also investigate the influence of parser selection for a biomedical event extraction downstream task,  and show that better intrinsic parsing performance does not always imply better extrinsic event extraction performance. Whether this pattern holds for other  information extraction tasks is left as future work.


\begin{backmatter}

\section*{Availability of data and material}

We make the retrained models available at \url{https://github.com/datquocnguyen/BioPosDep}.

\section*{Abbreviations}

NLP: Natural language processing; POS: Part-of-speech; SOTA: State-of-the-art; LAS: Labeled attachment score; UAS: Unlabeled attachment score; PTB: Penn treebank; WSJ:  Wall street journal; EPE: Extrinsic parser evaluation; CNN: Convolutional neural network; LSTM: Long short-term memory; BiLSTM: Bidirectional LSTM; CRF: Conditional random field; OOV: Out-of-vocabulary.

\section*{Competing interests}
  The authors declare that they have no competing interests.
  
\section*{Funding}
This work was supported by the ARC Discovery Project DP150101550 and ARC Linkage Project LP160101469. 

\section*{Author's contributions}
    DQN designed and conducted all the experiments, and drafted the manuscript. KV contributed to the manuscript and provided valuable comments on the design of the experiments.  Both authors have read and approved this manuscript.
    
\section*{Ethics approval and consent to participate}
Not applicable.

\section*{Consent for publication}
Not applicable.

\section*{Acknowledgement}
This research was also supported by use of the Nectar Research Cloud, a collaborative Australian research platform supported by the National Collaborative Research Infrastructure Strategy (NCRIS).


\bibliographystyle{bmc-mathphys} 
\bibliography{REFs}      

\end{backmatter}
\end{document}